\documentclass{INTERSPEECH2023}
\usepackage{tabularx}
\usepackage{multirow}

\interspeechcameraready

\title{Don't Stop Self-Supervision: Accent Adaptation of Speech Representations via Residual Adapters}
\name{Anshu Bhatia $^*{^1}{^\dag}$ \quad Sanchit Sinha $^*{^2}{^\ddag}$  \quad Saket Dingliwal $^1$ \quad Karthik Gopalakrishnan $^1$ \quad \\ 
Sravan Bodapati $^1$ \quad Katrin Kirchhoff $^1$}
\address{
  $^1$ AWS AI Labs \quad $^2$ University of Virginia}
\email{$^2$ss7mu@virginia.edu \\
$^1$\{anshubha, skdin, karthgop, sravanb, katrinki\}@amazon.com
}
\newcommand{\ssl}{$\mathcal{L}_{\text{SSL}}$ }
\newcommand{\ctc}{$\mathcal{L}_{\text{CTC}}$ }
\begin{document}

\maketitle
\begin{abstract}
Speech representations learned in a self-supervised fashion from massive unlabeled speech corpora have been adapted successfully toward several downstream tasks. However, such representations may be skewed toward canonical data characteristics of such corpora and perform poorly on atypical, non-native accented speaker populations. With the state-of-the-art HuBERT model as a baseline, we propose and investigate self-supervised adaptation of speech representations to such populations in a parameter-efficient way via training accent-specific residual adapters. We experiment with 4 accents and choose automatic speech recognition (ASR) as the downstream task of interest. We obtain strong word error rate reductions (WERR) over HuBERT-large for all 4 accents, with a mean WERR of 22.7\% with accent-specific adapters and a mean WERR of 25.1\% if the entire encoder is accent-adapted. While our experiments utilize HuBERT and ASR as the downstream task, our proposed approach is both model and task-agnostic.
\end{abstract}
\noindent\textbf{Index Terms}: speech recognition, residual adapters, accents, self-supervision, fairness 

\def\thefootnote{*}\footnotetext{~Equal contribution}
\def\thefootnote{\textdagger}\footnotetext{~Corresponding author}
\def\thefootnote{\ddag}\footnotetext{~Work done as an intern at AWS AI Labs}

\section{Introduction}
\label{section:intro}

Self-supervised learning has been a dominant paradigm in natural language processing (NLP) \cite{devlin2018bert} and in recent years, it has also been adopted by the speech community to learn high-fidelity representations \cite{hsu2021hubert, baevski2020wav2vec, chen2022wavlm, ling2020deep} that capture various non-lexical aspects of speech and audio such as lip-smacking, laughter, hesitation, etc. In this paradigm, the targets to learn are derived from the input signal itself, making the learned representations more powerful in principle compared to those learned using textual labels and annotations of any kind. These powerful base representations have been successfully adopted for several downstream tasks \cite{yang21c_interspeech}, some of which include: ASR, speaker identification and speech translation. Pre-training models with a very large number of parameters on proportionally large datasets has been a central theme in self-supervised learning. However, these datasets may understandably fall short in terms of sufficiently capturing non-canonical and diverse speech and audio characteristics such as rare non-native accents, stammering, etc. This leads to great disparity in downstream task performance across well-represented and underrepresented speaker populations. This data problem has also existed with supervised models for specific tasks such as ASR and in such scenarios, the typical path has been to patch task performance by collecting task-specific labeled datasets with non-canonical characteristics and fine-tuning for the task \cite{tomanek-etal-2021-residual}. This unfortunately entangles speech and audio characteristics with the task itself, which can limit effective learning of such characteristics in task-specific representations as well as limiting their re-usability across tasks.

In this paper, we consequently posit that continued self-supervised learning of speech and audio representations on task-agnostic unlabeled datasets is an effective strategy to adapt to non-canonical speech characteristics. The specific characteristic we choose to study is accents but the methodology holds for any characteristic. We propose learning different high-dimensional spaces for different accents via independently adding residual adapters for each target accent to the model and continuing pre-training on accent-specific datasets. Since residual adapters are parameter-wise much smaller than the base model, this serves as a parameter-efficient way for personalized adaptation without over-fitting and saves on storage costs for inference since only a single copy of the base model needs to be stored. We conduct our experiments with HuBERT-large \cite{hsu2021hubert} as the base model and ASR as the downstream task but posit that our proposed approach is both model and task agnostic. Our chosen base model is a state-of-the-art model with low word error rates on canonical datasets such as LibriSpeech. By design, we pick 4 non-native English accents where the HuBERT-large model has high word error rates (WER), in the range 24-50\% and show strong results on all 4 accents with over 22\% WERR over the baseline. Previous work has shown improvements in WER on such accents by supervised training using labeled datasets \cite{tomanek-etal-2021-residual}. In contrast, we achieve our WER improvements by continuing to self-supervise models using unlabeled data alone. We show that the gains from adapting to an accent using a particular dataset translate to other evaluation sets with the same accent as well, indicating that the effectiveness of our approach is due to adaptation to the accents' acoustic characteristics and not other confounding factors. Finally, we also explore the degree of parameter-efficiency possible when adapting to target accents, finding that we can achieve strong WERR over the baseline while updating only 16\% of the total model parameters. 

\begin{figure*}[t]
  \centering
  \includegraphics[width=\textwidth]{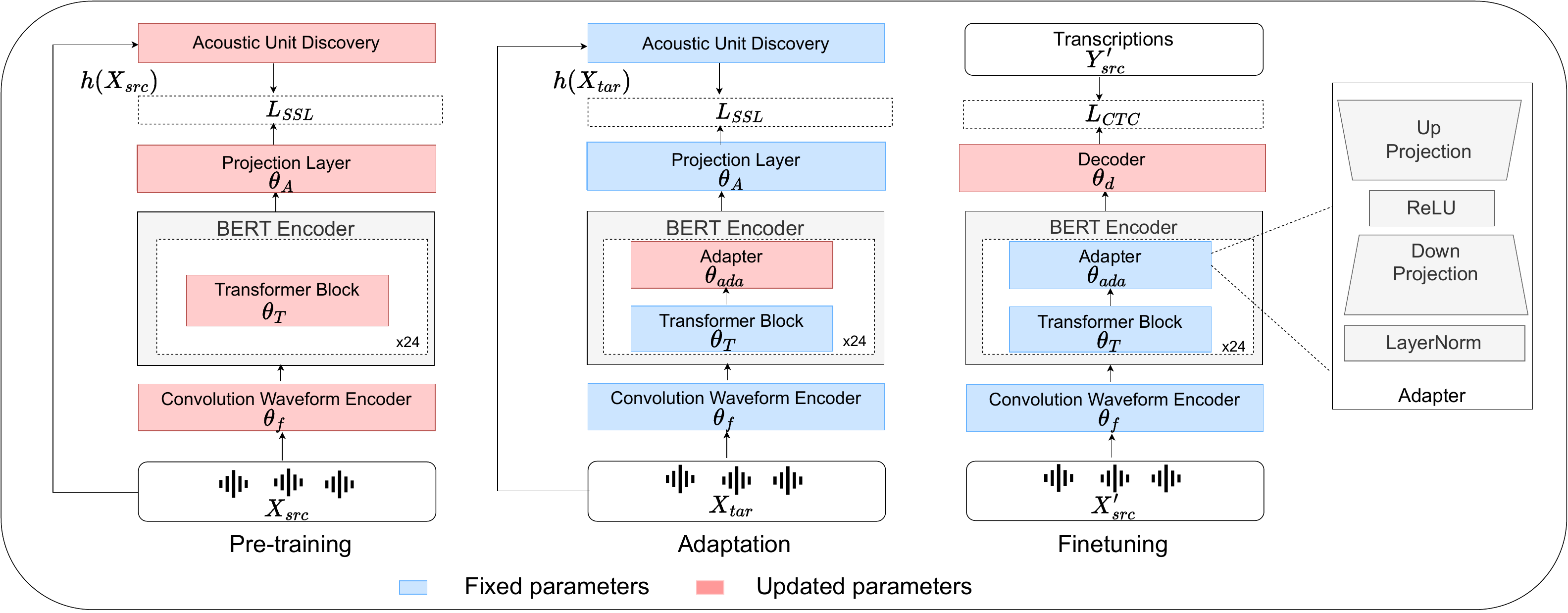}
  \caption{Accent-Adaptive Continual Self-Supervision: Three training stages to improve the performance of any Transformer-based speech foundation model on any downstream task for a speaker group.}
  \label{fig:model_training_stages}
\end{figure*}

\section{Methodology}
\label{section:method}
In this work, we propose a model-agnostic as well as a task-agnostic method to adapt audio representations for speakers of a particular group. It can be leveraged to improve the performance of any Transformer-based speech model in the literature \cite{hsu2021hubert,baevski2020wav2vec, radford2022robust} on any downstream speech task. We showcase the efficacy of our approach on the widely used HuBERT model \cite{hsu2021hubert} with state-of-the-art performance on different speech tasks \cite{yang21c_interspeech} at the time of our experiments. We evaluate our accent-specific audio representations on one of the important tasks, i.e., ASR.

\subsection{Background}
The HuBERT model consists of a convolutional waveform encoder \cite{baevski2020wav2vec}, a BERT encoder \cite{devlin2018bert} and a projection layer. The convolutional waveform encoder (parameterized by $\theta_f$), takes audio as an input to generate a feature sequence at a 20ms duration. The BERT encoder consists of $N$ identical Transformer blocks (with parameters $\theta_T$), stacked one after the other. It takes the input from the waveform encoder and passes the output feature sequence (768 dimensional vector sequence) to the projection layer. The projection layer with parameters $\theta_A$, maps the feature sequence to the target sequence. These frame-level targets are provided by an independent clustering model like k-means and is called acoustic unit discovery module. Let $X = [x_1 \cdot \cdot \cdot x_T]$ denote a speech utterance of $T$ frames. The discovered hidden units are denoted with $h(X) = Z = [z_1 \cdot \cdot \cdot z_T ]$, where $z_t \in [C]$ is a $C$-class categorical variable and $h$ is a clustering model. As defined in \cite{hsu2021hubert}, the probability of predicting  $c^{\text{th}}$ cluster center at time step $t$ by the HuBERT model ($\Theta = \{\theta_f, \theta_T, \theta_A\}$) is denoted by $p_{\Theta}(c | X, t)$.

The HuBERT model is trained in two stages where the first stage is self-supervision with unlabeled audio sequences while the second stage involves fine-tuning on a downstream task using labeled data. During the pre-training of the model, a subset of indices ($M \subset [T]$) are masked to create $\tilde{X} = r(X, M)$ denoting a corrupted version of $X$ where $x_t$ is replaced with a mask embedding $\tilde{x}$ if $t \in M$. The self-supervision loss (\ssl) is defined as the cross-entropy loss in predicting the targets for the masked time-steps of an audio sequence. 
\begin{equation}
    \mathcal{L}_{\text{SSL}} (X, r, \Theta, h, M) = \sum_{t \in M} \log p_{\Theta}(z_t | \tilde{X}, t)
\label{eq:ssl}
\end{equation}

After pre-training, the projection layer is removed and a light-weight decoder is used to map the audio representations from the BERT encoder to the output of the downstream task at hand. For ASR, it is used to predict targets from a pre-defined vocabulary (26 English characters, a space token, an apostrophe, and a special CTC blank symbol). Our decoder architecture is similar to \cite{yang21c_interspeech}, where the output vector sequence from each of the $N$ Transformer blocks in the BERT encoder is multiplied with scalar weights ($W = [w_1 \cdot \cdot w_{N}]$), added and then passed through a vanilla 2-layer 1024-unit bidirectional LSTM \cite{hochreiter1997long} (with parameters $\theta_{\text{LSTM}}$), which is used to predict the output sequence. Let $Y = [y_1 \cdot \cdot y_{T^{'}}]$ denote the ground truth labels. The parameters of the decoder ($\theta_d = \{W, \theta_{\text{LSTM}}\}$) are learned by minimizing the connectionist temporal classification (CTC) \cite{graves2006connectionist} loss \ctc $(X,Y,\Theta,\theta_d) = \log p_{\Theta, \theta_d}(Y|X)$ between the predicted sequence and the ground truth. During this stage, all the parameters ($\Theta$) in the HuBERT model are frozen. As shown by \cite{yang21c_interspeech}, this helps us save compute and storage costs with little to no degradation in downstream task performance as it allows for using a common encoder model for different tasks. 

\subsection{Accent-Adaptive Continual Self-Supervision}
Our approach for generating accent-specific audio representations is simple and effective. It can be used to improve performance of any downstream speech task. We simply introduce an additional training stage where we use unlabeled audio from our atypical target accent for continuing self-supervised pre-training. As shown in Fig. \ref{fig:model_training_stages}, we train our model for a task in three stages. Let $X_{src}$, $X_{tar}$ represents the unlabeled audio sequences from the generic data and the target accent respectively. We will denote the generic labeled data for a particular task like ASR with $\{X^{'}_{src}, Y^{'}_{src}\}$ which may or may not overlap with $X_{src}$. The first stage is same as defined in the previous subsection, where we use generic unlabeled data $X_{src}$ to minimize SSL loss defined in Eq. \ref{eq:ssl}. In the second stage, we continue to minimize the same loss but with $X_{tar}$. Finally in the third stage, we learn the task-specific decoder parameters ($\theta_d$) by minimizing \ctc $(\{X'_{src}, Y'_{src}\}, \Theta, \theta_d)$. This additional self-supervision helps to modify the generic audio representations to capture the acoustic features relevant to the target accent. These accent-specific representations can improve performance of the model on any downstream task for the speaker group with a particular target accent.

Although continued self-supervision improves performance, it comes at the computational and memory costs of training, storing and deploying separate BERT encoder models for different accents. To overcome these additional costs, we introduce parameter-efficiency using residual adapters \cite{houlsby2019parameter, pfeiffer2020adapterhub}. Adapters were first introduced for Transformer-based language models to adapt these large models to different tasks. With a handful of additional parameters per task, adapters have been shown to influence the output from the Transformer and hence make them task-specific \cite{houlsby2019parameter, dingliwal2023personalization}. In our work, we extend the application of adapters to speech where they are used for adapting audio representations for different accents. We introduce an adapter sandwiched between every Transformer block of the BERT encoder of the HuBERT model. Each adapter module consists of a layer normalization, a feed-forward network to project the vector sequence from the Transformer block to a new bottle-neck dimension $B_{ada}$, ReLU activation and finally another feed-forward network to project back the vector sequence to the original dimension. The output from the adapter is added back to the original vector sequence and fed to the next Transformer block. We collectively denote parameters of all the adapters in our model by $\theta_{ada}$. In our accent-adaptive self-supervision stage, rather than updating all the parameters of the HuBERT model ($\Theta$), we only update $\theta_{ada}$ keeping $\Theta$ constant. This ensures that we can still obtain accent-specific audio representations while storing a much smaller set of accent-specific parameters relative to HuBERT for each target accent.

Prior work \cite{tomanek-etal-2021-residual, fan2022draft}  introduced accent-specific adapters in speech models by learning accent information from labeled data for downstream tasks. However, in sharp contrast, we hypothesize that the accent information can be efficiently captured without using any labels from the downstream task. Accent of a speaker is a speech characteristic and the self-supervised objective of predicting the masked audio sequence targets is suitable to capture accent-specific information. Our learned audio representations for a target accent are more general and efficient than the prior work as they can used for any downstream speech task and they do not require any additional labels per accent.

\section{Experimental Setup}
\subsection{Datasets}
For our experiments, we use the publicly available version of 60K hours of LibriLight \cite{kahn2020libri} as the generic unlabeled data ($X_{src}$) in the pre-training stage. Similarly, we use 960 hours of paired speech-text data from LibriSpeech \cite{7178964} for the fine-tuning stage. This is representative of the standard data setting used by many SSL models \cite{hsu2021hubert, baevski2020wav2vec, chen2022wavlm}. We verify the claims of our methodology by adapting our models on four different target accents. The unlabeled data ($X_{tar}$) for two of these four accents, i.e., Indian (\texttt{in}) and Scottish (\texttt{sc}), is taken from Mozilla Common Voice Corpus v6.1 (MCV) \cite{ardila2019common} dataset, while we collect the audio sequences for the other two accents,  German (\texttt{de}) and Chinese (\texttt{cn}) in-house. These audio sequences are conversational in nature and are collected by making diverse set of speakers of a particular accent read dialogues. The details of the number of utterances and hours of recordings used for training, validation and evaluation are shown in Table \ref{tab:Training data distribution}. For three of the four accents, we use 30 hours of unlabeled audio while we only use 6.6 hours of \texttt{sc} accent. This is significantly smaller than 60K hours of audio used for pre-training in the first stage. Note that the ground truth labels of any of the accent-specific training and validation datasets are never used in our experiments. To test generalization of our accent-adapted models in different settings, we use additional evaluation datasets. We separately collect 10.1 hours of paired Conversational (Conv.) audio and text from speakers with Indian accents. We also use 2 hours of Indian speaker-specific audio and text from publicly available VoxForge \cite{Voxforge.org} dataset for evaluation.
\begin{table}
\caption{Data distribution (utterance count and total hours) for target accents used in our experiments. We use Indian (\texttt{in}), Scottish (\texttt{sc}), German (\texttt{de}) and Chinese (\texttt{cn}) as target accents.}
\label{tab:Training data distribution}
\resizebox{\columnwidth}{!}{
\begin{tabular}{|c|c|c|c|c|c|c|c|c|}
\hline
\multirow{2}{*}{Data} & \multicolumn{2}{c|}{\texttt{in}} & %
    \multicolumn{2}{c|}{\texttt{sc}} & %
    \multicolumn{2}{c|}{\texttt{de}} & %
    \multicolumn{2}{c|}{\texttt{cn}} \\
\cline{2-9} &
 Utt & Hrs & Utt & Hrs & Utt & Hrs & Utt & Hrs \\
\hline
train & 19699 & 31.4 & 3709 & 6.6 &15854  & 30.2 & 16731 & 29.3  \\
\hline
valid & 6567 & 10.5 & 1237 & 2.2 & 5365 & 10.16 & 5783 & 10.56  \\
\hline
test & 6517 & 10.5 & 1237  & 2.2 & 5216  & 9.67  & 4000 &  9.83 \\
\hline
\end{tabular}}
\end{table}

\subsection{Model settings} 
As defined in Section \ref{section:method}, we use HuBERT-large model with $N=24$ Transformer blocks. For the first stage of generic pre-training, the HuBERT model parameters ($\Theta$), the acoustic unit discovery module ($h$), the mask indices ($M$) and the masking function ($r$) are obtained from their open-sourced versions by \textit{fairseq} \cite{ott2019fairseq}. This model is referred to as \textit{baseline} in our tables. They use 60K hours of unlabeled LibriLight data \cite{kahn2020libri} and minimize \ssl to obtain the model parameters. Further, their clustering model ($h$) is a k-means model where cluster centers are identified using the output feature sequence from the $9^{th}$ BERT encoder layer of the pre-trained HuBERT-base model. For the next stage of accent-adaptive self-supervision, we use the same clustering model $h$ to obtain targets for accented-speech i.e. $Z_{tar} = h(X_{tar})$. In this stage, we train the model with and without the adapters i.e., \textit{Accent-Adapters} and \textit{Accent-HuBERT} respectively. For the model without the adapters, we update all the parameters of the HuBERT model ($\Theta$) with a learning rate of $2e-5$ and linear warmup phase of 20k updates. The maximum number of tokens in each batch is set to 300k and the model is trained for 150k steps and finally the best model is chosen using the self-supervision loss value on the unlabeled accent-specific validation dataset. When using adapters, all settings are the same except that we freeze $\Theta$ and only update $\theta_{ada}$ using a learning rate scheduler with peak learning rate of $1e-3$, a linear warmup phase of 75k steps, followed by polynomial decay till 0. For our final stage of task-specific fine-tuning, we use the same experimental settings as \textit{s3prl} \cite{yang21c_interspeech} for ASR. For all three models, we train the decoder parameters ($\theta_d$) with 16 batch size and $5.0e-5$ learning rate till the decrease in the training loss between subsequent epochs is less than a certain threshold. Similar to \cite{yang21c_interspeech}, we use the LibriSpeech official 4-gram language model powered by KenLM \cite{heafield2011kenlm} and flashlight toolkit \cite{pratap2019wav2letter++} fused together with our models during decoding. 
\section{Results}

For our experiments, the baseline is the state-of-the-art HuBERT model \cite{hsu2021hubert} that achieves a WER of 2.3 and 4.6 on \textit{test-clean} and \textit{test-other} subsets of LibriSpeech \cite{7178964} in a similar setting as used in \cite{yang21c_interspeech}. As highlighted previously in Section \ref{section:intro}, we specifically aim to improve the performance of the baseline model for the speakers of the accents that see high WERs even though the model performs well on the standard benchmarks. For example, the WER of the baseline model on \texttt{in} and \texttt{sc} accent from the publicly available MCV dataset are 24.8 and 52.0 respectively. All the numbers reported in our tables are WER Reduction \% (WERR) over the baseline model. Important findings from our experiments are summarized below:

\begin{table}[th]
\caption{\textbf{WERR (\%)} of accent-specific models as compared to the baseline. The baseline model is HuBERT-large, a state-of-the-art model that performs poorly on the selected target accents from the publicly available MCV datasets. The baseline model has a \textbf {WER of 24.8 and 52.0} on \texttt{in} and \texttt{sc} accents respectively.}
\label{tab:WER on different models}
\centering
\begin{tabular}{|l|c|c|c|c|}
\hline
   \multirow{2}{*}{\textbf{Model}}  & \multicolumn{4}{c|}{\textbf{WERR (\%) on Target Accents}} \\%
    \cline{2-5}
    & \texttt{in} & \texttt{sc} & \texttt{de} & \texttt{cn}   \\
\hline
\textit{HuBERT-large}   & -- & -- & --  & --   \\
\hline
 \textit{Accent-Adapters}   & 23.9\% & 28.5\% & 17.7\%  & 20.8\%   \\
 \hline
\textit{Accent-HuBERT} & 27.2\% & 27.8\% & 22.8\%  & 22.7\%    \\
 \hline
\end{tabular}
\end{table}

\noindent \textbf{Continued self-supervision enables learning rich task-agnostic representations for different accents}: We showcase that models with continued self-supervision perform significantly better than the baseline on the ASR task. In Table \ref{tab:WER on different models}, our models reduce WERs on all four accents without using any accent-specific labeled data during training. \textit{Accent-Adapters} and \textit{Accent-HuBERT} achieve 22.7\% and 25.1\% WERR on average respectively. Since we use the same task-specific fine-tuning setting for both the baseline and our methods, we attribute the improvements to richer audio representations learned by the base model that can adapt to speech characteristics related to the target accent. This is an important finding as it enables performance gains on any downstream task without spending resources on collection of task-specific labeled data.

To validate our hypothesis that the improvement in ASR performance is indeed the result of richer acoustic representations of the accented speech, we evaluate our models on unseen datasets. We use the model trained on the audio from the \texttt{in} accent in the MCV dataset and evaluate it on two independently collected datasets by speakers of the same accent.
The WERR \%ages from the baseline are summarized in Table \ref{tab:Different in accent datasets}. We see a significant 12.6\% and 6.7\% reduction in WER of our \textit{Accent-HuBERT} model on the \texttt{in} accent subset of VoxForge and Conversational data respectively. 
We attribute these improvements specifically to accent-related acoustic features learned by the HuBERT model as it is the only common factor between the training and the evaluation dataset. All other confounding factors related to the unlabeled audio like content of the audio, individual speaker related features, signal-noise ratio etc., are factored out in these evaluations. This showcases the robustness of the improvements stemming from our methodology.  

\begin{table}[th]
\caption{\textbf{WERR (\%)} over the the baseline of the accent-specific models trained on \texttt{in} accent of the MCV dataset when evaluated on the subsets of VoxForge and Conversational datasets with the same accent. The baseline model has a \textbf{WER of 18.6} on this subset from the VoxForge dataset.}
\label{tab:Different in accent datasets}
\centering
\begin{tabular}{|c|c|c|}
\hline
\multirow{3}{*}{\textbf{Model}}  & %
     \multicolumn{2}{c|}{\textbf{WERR (\%) on Datasets}} \\%
\cline{2-3} 
   &  \texttt{VoxForge} & \texttt{Conv.}   \\
\hline
\textit{HuBERT-large}   & -- & --     \\
 \hline
\textit{Accent-Adapters}   & 12.6\% & 6.7\%     \\
 \hline
\textit{Accent-HuBERT}  & 19.5\% & 8.9\%  \\
 \hline
\end{tabular}
\end{table}

\noindent \textbf{Adapters are a cost-effective way to capture accent-specific features in large self-supervised speech models}: 
Our baseline model HuBERT-large ($\Theta$) has 317M parameters. Fine-tuning, storing and deploying such models individually for each speaker group can be limited by computational and memory constraints, although that would give the best performance in principle. Adapters, on the other hand, can achieve similar performance using $\sim$85\% less parameters per speaker group. Our findings are in line with many prior works in natural language processing (NLP) \cite{houlsby2019parameter, pfeiffer2020adapterhub} and speech \cite{tomanek-etal-2021-residual, fan2022draft}, where adapter modules have been showcased to influence the output of the Transformer model using bottle-neck layers. The dimension of this bottle-neck layer ($B_{ada}$) is used to trade-off between the performance and cost of the model. In Table \ref{tab:Adapter size ablations}, we provide an ablation for the choice of $B_{ada}$ and the WERR \% on one of the accents used for evaluation i.e., \texttt{in} from the MCV test set. With just 16\% of the base model parameters, we see a strong 23.9\% WERR over the baseline. We see diminishing returns of performance improvement as we increase the size of the bottle-neck dimension beyond 1024. Therefore, $B_{ada} = 1024$ was the choice for all the other experiments in this work.
\begin{table}[th]
\caption{\textbf{WERR (\%)} over the baseline and the accent-specific parameter count (as \%age of the count of the base model parameters) of different accent-specific models with different bottleneck sizes ($B_{ada}$). These reductions are on top of the baseline \textbf{WER of 24.8 on} \texttt{in} accent from the MCV dataset.}
\label{tab:Adapter size ablations}
\centering
\begin{tabular}{|l|c|c|c|}
 \hline
  \textbf{Model} & $\mathbf{B_{ada}}$ & \textbf{\% params}  & 
  \textbf{WERR (\%)} \\
  \hline
  \textit{\textit{HuBERT-large}} & NA & -- & -- \\
  \hline
 \multirow{3}{*}{\textit{Accent-Adapter}} & 512 &  8  & 19.8 \\
                                          &1024 & 16  & 23.9 \\
                                          &2048 & 32  & 24 \\
  \hline
  \textit{Accent-HuBERT} & NA & 100 & 27.2 \\
  \hline
\end{tabular}
\end{table}
\section{Conclusions}
In this paper, we propose adapting self-supervised speech representations to atypical accents by continuing to perform self-supervision using such data. To the best of our knowledge, we are the first to show strong improvements over state-of-the-art baselines by adapting models using self-supervision on unlabeled accented data. We experiment with modifying the base encoder by adding adapters to each Transformer block and updating the adapters alone during accent-adaptive pre-training, as well as with updating the entire encoder during accent-adaptive pre-training. Our method achieves strong WERR over the state-of-the-art on 4 different non-native accents. We achieve an average 22.7\% WERR  when using adapters and an average of 25.1\% WERR when updating the entire encoder. We also show that our models adapted to an accent using a given dataset perform well on other evaluation sets with similar speaker characteristics, thus validating our hypothesis that our models adapt by learning accent-specific acoustic representations from the target speech. Our approach is parameter-efficient and we show strong WERR by updating just 16\% of the model parameters. Although, we conduct our experiments with ASR as the downstream task in this work, we posit that our approach is task agnostic, since we perform adaptation during the pre-training stage.

Our proposed approach has great practical viability due to 2 reasons: (a) we can adapt using unlabeled data alone, which is far easier and cheaper to obtain compared to high-quality labeled data, and (b) we can adapt models to different accents in a parameter-efficient way with only a small number of accent-specific parameters, without needing to incur the memory and compute costs of maintaining large models for each accent. While our current work focuses on adapting to unlabeled accented data, effectively utilizing a small amount of labeled accented data alongside accent-adaptive self-supervision is a promising future direction to explore.

\bibliographystyle{IEEEtran}

\begin{thebibliography}{10}
\providecommand{\url}[1]{#1}
\csname url@samestyle\endcsname
\providecommand{\newblock}{\relax}
\providecommand{\bibinfo}[2]{#2}
\providecommand{\BIBentrySTDinterwordspacing}{\spaceskip=0pt\relax}
\providecommand{\BIBentryALTinterwordstretchfactor}{4}
\providecommand{\BIBentryALTinterwordspacing}{\spaceskip=\fontdimen2\font plus
\BIBentryALTinterwordstretchfactor\fontdimen3\font minus
  \fontdimen4\font\relax}
\providecommand{\BIBforeignlanguage}[2]{{%
\expandafter\ifx\csname l@#1\endcsname\relax
\typeout{** WARNING: IEEEtran.bst: No hyphenation pattern has been}%
\typeout{** loaded for the language `#1'. Using the pattern for}%
\typeout{** the default language instead.}%
\else
\language=\csname l@#1\endcsname
\fi
#2}}
\providecommand{\BIBdecl}{\relax}
\BIBdecl

\bibitem{devlin2018bert}
J.~Devlin, M.-W. Chang, K.~Lee, and K.~Toutanova, ``{BERT: Pre-training of Deep
  Bidirectional Transformers for Language Understanding},'' \emph{arXiv
  preprint arXiv:1810.04805}, 2018.

\bibitem{hsu2021hubert}
W.-N. Hsu, B.~Bolte, Y.-H.~H. Tsai, K.~Lakhotia, R.~Salakhutdinov, and
  A.~Mohamed, ``{HuBERT: Self-Supervised Speech Representation Learning by
  Masked Prediction of Hidden Units},'' \emph{IEEE/ACM Transactions on Audio,
  Speech, and Language Processing}, vol.~29, pp. 3451--3460, 2021.

\bibitem{baevski2020wav2vec}
A.~Baevski, Y.~Zhou, A.~Mohamed, and M.~Auli, ``{wav2vec 2.0: A Framework for
  Self-Supervised Learning of Speech Representations},'' \emph{Advances in
  neural information processing systems}, vol.~33, pp. 12\,449--12\,460, 2020.

\bibitem{chen2022wavlm}
S.~Chen, C.~Wang, Z.~Chen, Y.~Wu, S.~Liu, Z.~Chen, J.~Li, N.~Kanda,
  T.~Yoshioka, X.~Xiao \emph{et~al.}, ``{WavLM: Large-Scale Self-Supervised
  Pre-Training for Full Stack Speech Processing},'' \emph{IEEE Journal of
  Selected Topics in Signal Processing}, vol.~16, no.~6, pp. 1505--1518, 2022.

\bibitem{ling2020deep}
S.~Ling, Y.~Liu, J.~Salazar, and K.~Kirchhoff, ``{Deep contextualized acoustic
  representations for semi-supervised speech recognition},'' in \emph{ICASSP
  2020-2020 IEEE International Conference on Acoustics, Speech and Signal
  Processing (ICASSP)}.\hskip 1em plus 0.5em minus 0.4em\relax IEEE, 2020, pp.
  6429--6433.

\bibitem{yang21c_interspeech}
S.~wen Yang, P.-H. Chi, Y.-S. Chuang, C.-I.~J. Lai, K.~Lakhotia, Y.~Y. Lin,
  A.~T. Liu, J.~Shi, X.~Chang, G.-T. Lin, T.-H. Huang, W.-C. Tseng, K.~tik Lee,
  D.-R. Liu, Z.~Huang, S.~Dong, S.-W. Li, S.~Watanabe, A.~Mohamed, and
  H.~yi~Lee, ``{SUPERB: Speech Processing Universal PERformance Benchmark},''
  in \emph{Proc. Interspeech 2021}, 2021, pp. 1194--1198.

\bibitem{tomanek-etal-2021-residual}
\BIBentryALTinterwordspacing
K.~Tomanek, V.~Zayats, D.~Padfield, K.~Vaillancourt, and F.~Biadsy, ``Residual
  adapters for parameter-efficient asr adaptation to atypical and accented
  speech,'' in \emph{Proceedings of the 2021 Conference on Empirical Methods in
  Natural Language Processing}.\hskip 1em plus 0.5em minus 0.4em\relax Online
  and Punta Cana, Dominican Republic: Association for Computational
  Linguistics, Nov. 2021, pp. 6751--6760. [Online]. Available:
  \url{https://aclanthology.org/2021.emnlp-main.541}
\BIBentrySTDinterwordspacing

\bibitem{radford2022robust}
A.~Radford, J.~W. Kim, T.~Xu, G.~Brockman, C.~McLeavey, and I.~Sutskever,
  ``{Robust Speech Recognition via Large-Scale Weak Supervision},'' \emph{arXiv
  preprint arXiv:2212.04356}, 2022.

\bibitem{hochreiter1997long}
S.~Hochreiter and J.~Schmidhuber, ``Long short-term memory,'' \emph{Neural
  computation}, vol.~9, no.~8, pp. 1735--1780, 1997.

\bibitem{graves2006connectionist}
A.~Graves, S.~Fern{\'a}ndez, F.~Gomez, and J.~Schmidhuber, ``{Connectionist
  Temporal Classification: labelling unsegmented sequence data with recurrent
  neural networks},'' in \emph{Proceedings of the 23rd international conference
  on Machine learning}, 2006, pp. 369--376.

\bibitem{houlsby2019parameter}
N.~Houlsby, A.~Giurgiu, S.~Jastrzebski, B.~Morrone, Q.~De~Laroussilhe,
  A.~Gesmundo, M.~Attariyan, and S.~Gelly, ``{Parameter-Efficient Transfer
  Learning for NLP},'' in \emph{International Conference on Machine
  Learning}.\hskip 1em plus 0.5em minus 0.4em\relax PMLR, 2019, pp. 2790--2799.

\bibitem{pfeiffer2020adapterhub}
J.~Pfeiffer, A.~R{\"u}ckl{\'e}, C.~Poth, A.~Kamath, I.~Vuli{\'c}, S.~Ruder,
  K.~Cho, and I.~Gurevych, ``{AdapterHub: A Framework for Adapting
  Transformers},'' in \emph{Proceedings of the 2020 Conference on Empirical
  Methods in Natural Language Processing: System Demonstrations}, 2020, pp.
  46--54.

\bibitem{dingliwal2023personalization}
S.~Dingliwal, M.~Sunkara, S.~Ronanki, J.~Farris, K.~Kirchhoff, and S.~Bodapati,
  ``{Personalization of CTC speech recognition models},'' in \emph{2022 IEEE
  Spoken Language Technology Workshop (SLT)}.\hskip 1em plus 0.5em minus
  0.4em\relax IEEE, 2023, pp. 302--309.

\bibitem{fan2022draft}
R.~Fan and A.~Alwan, ``{DRAFT: A Novel Framework to Reduce Domain Shifting in
  Self-supervised Learning and Its Application to Children's ASR},''
  \emph{arXiv preprint arXiv:2206.07931}, 2022.

\bibitem{kahn2020libri}
J.~Kahn, M.~Riviere, W.~Zheng, E.~Kharitonov, Q.~Xu, P.-E. Mazar{\'e},
  J.~Karadayi, V.~Liptchinsky, R.~Collobert, C.~Fuegen \emph{et~al.},
  ``{Libri-Light: A Benchmark for ASR with Limited or No Supervision},'' in
  \emph{ICASSP 2020-2020 IEEE International Conference on Acoustics, Speech and
  Signal Processing (ICASSP)}.\hskip 1em plus 0.5em minus 0.4em\relax IEEE,
  2020, pp. 7669--7673.

\bibitem{7178964}
V.~Panayotov, G.~Chen, D.~Povey, and S.~Khudanpur, ``{Librispeech: An ASR
  corpus based on public domain audio books},'' in \emph{2015 IEEE
  International Conference on Acoustics, Speech and Signal Processing
  (ICASSP)}, 2015, pp. 5206--5210.

\bibitem{ardila2019common}
R.~Ardila, M.~Branson, K.~Davis, M.~Henretty, M.~Kohler, J.~Meyer, R.~Morais,
  L.~Saunders, F.~M. Tyers, and G.~Weber, ``{Common Voice: A
  Massively-Multilingual Speech Corpus},'' \emph{arXiv preprint
  arXiv:1912.06670}, 2019.

\bibitem{Voxforge.org}
Voxforge.org, ``Free speech... recognition (linux, windows and mac) -
  voxforge.org,'' \url{http://www.voxforge.org/}, accessed 07/25/2022.

\bibitem{ott2019fairseq}
M.~Ott, S.~Edunov, A.~Baevski, A.~Fan, S.~Gross, N.~Ng, D.~Grangier, and
  M.~Auli, ``{fairseq: A Fast, Extensible Toolkit for Sequence Modeling},'' in
  \emph{Proceedings of NAACL-HLT 2019: Demonstrations}, 2019.

\bibitem{heafield2011kenlm}
K.~Heafield, ``{KenLM: Faster and Smaller Language Model Queries},'' in
  \emph{Proceedings of the sixth workshop on statistical machine translation},
  2011, pp. 187--197.

\bibitem{pratap2019wav2letter++}
V.~Pratap, A.~Hannun, Q.~Xu, J.~Cai, J.~Kahn, G.~Synnaeve, V.~Liptchinsky, and
  R.~Collobert, ``{Wav2letter++: A fast open-source speech recognition
  system},'' in \emph{ICASSP 2019-2019 IEEE International Conference on
  Acoustics, Speech and Signal Processing (ICASSP)}.\hskip 1em plus 0.5em minus
  0.4em\relax IEEE, 2019, pp. 6460--6464.

\end{thebibliography}

\end{document}